\documentclass[runningheads]{llncs}
\usepackage[T1]{fontenc}
\usepackage{graphicx}
\usepackage{amsmath} 
\usepackage{multirow}
\usepackage{booktabs}
\usepackage{amssymb}
\usepackage{xcolor}
\begin{document}
\title{Product units in gated recurrent units improve nuclear-mass prediction}
%Predicting Nuclear Mass Excess via Additive-Multiplicative Product-Unit Gated Recurrent Units}
\author{Ziyuan Li\inst{1,2} \and
Paulo S.A. Freitas \inst{3} \and
John W. Clark \inst{3,4} \and
Babette Dellen \inst{1}
}
\authorrunning{Li et al.}
\institute{Department of Mathematics, Informatics and Technology, University of Applied Sciences Koblenz, Joseph-Rovan-Allee 2, 53424 Remagen, Germany\\ \email{dellen@hs-koblenz.de} \and
Technical University of Munich, Munich, Germany \and
Department of Mathematics, University of Madeira, Campus Universitário da Penteada, Funchal, 9020-105, Portugal \and
Department of Physics, Washington University in St. Louis, 1 Brookings Drive, St. Louis, 63130, MO, USA
}
\maketitle              % typeset the header of the contribution
\begin{abstract}
The prediction of masses of atomic nuclei using machine learning can complement theoretical models and advance the exploration of poorly known domains of the nuclear chart. We propose a machine learning technique based on gated recurrent units (GRU), which have demonstrated competitive performance in nuclear-mass prediction by exploiting long-term dependencies. By integrating multiplicative interactions and product-unit transformations within recurrent units, we report significant improvements in nuclear-mass prediction. Computations are performed in the complex domain to jointly capture amplitude and phase dynamics. For interpolation and temporal-extrapolation tasks based on the atomic mass evaluation (AME2016 and AME2020), the complex additive-multiplicative product-unit gated recurrent unit (AM-PU-GRU) model consistently achieves the lowest prediction errors, with an interpolation RMSE of 0.227 $\pm$ 0.004 MeV and an extrapolation RMSE of 0.179 $\pm$ 0.015 MeV. These results surpass other state-of-the-art machine learning models and also outperform the real-valued GRU baseline and product-unit ablation variants, while remaining robust to different theoretical priors, including WS4 and SEMF. Our findings establish complex-valued product-unit recurrent networks as a new benchmark for sequence-based nuclear-mass prediction.

\end{abstract}
\section{Introduction}
Accurate prediction of nuclear masses is a fundamental problem in nuclear physics, with important implications for nuclear--structure theory, nucleosynthesis pathways, and applications in nuclear energy and astrophysics. Experimental evaluations such as AME2016 \cite{kondev2017ame20161,kondev2017ame20162} and AME2020 \cite{huang2021ame,wang2021ame} provide precise measurements for many nuclides, yet large regions of the nuclear chart remain inaccessible. This motivates theoretical and data-driven approaches capable of both interpolation in known regions and extrapolation to unknown nuclides.

Traditional mass models, such as the Weizsäcker--Skyrme Model (Version 4) (WS4) \cite{wang2014surface}, achieve high accuracy through physics-motivated parametrizations, but their predictive power in unexplored regions is limited \cite{jalili2025deep}. In parallel, machine learning methods have emerged to complement nuclear-mass prediction, notably recurrent neural network (RNN) \cite{jalili2025deep}, gated recurrent unit (GRU) \cite{cho2014learning,jalili2025deep}, and mixture density network (MDN) \cite{mumpower2022physically,mumpower2023bayesian,li2024atomic} have achieved strong interpolation accuracy within experimentally known regions, sometimes comparable to physics-based models. These approaches benefit from their ability to learn complex nonlinear relationships directly from data and have been shown to reduce prediction errors when sufficient experimental measurements are available. However, many of these architectures remain limited in extrapolation capability, as they primarily rely on additive representations and struggle to capture the higher-order nonlinear dependencies inherent in nuclear-mass systematics.

The product-unit (PU) \cite{durbin1989product,leerink1994learning,dellen2019function,dellen2024predicting,li2024data,li2025deep,li2025advancing} approach to machine learning was introduced as an alternative to summation-based formulations, enabling multiplicative interactions that provide compact representations of polynomial and power-law relations. More recently, complex-valued PU extensions \cite{dellen2024predicting,li2024data} have been explored, enabling joint modeling of amplitude and phase dynamics.

Building on these advances, we propose complex-valued GRU extensions for nuclear mass prediction. By integrating multiplicative interactions and product-unit transformations into recurrent frameworks, we obtain two novel architectures, the multiplicative-interaction product-unit GRU (MI-PU-GRU) and the additive-multiplicative product-unit GRU (AM-PU-GRU).

We next evaluate these architectures on interpolation and temporal extrapolation based on AME2016 and AME2020, comparing against real-valued baselines, ablations, and prior-informed setups. The complex AM-PU-GRU model achieves the lowest error on both tasks, setting a new benchmark for sequence-based nuclear-mass prediction.

\section{Related Work}
Product units (PUs) were introduced as an alternative to summation-based neurons in neural networks. Instead of computing a weighted sum followed by a nonlinear activation, a PU models multiplicative interactions by raising each input to a learnable exponent and taking their product according to
\begin{equation}
y = \prod_{i=1}^n x_i^{w_i} = \exp\left( \sum_{i=1}^n w_i \log |x_i| \right).
\label{eq:pu_real}
\end{equation}
This formulation substantially enhances the expressive capacity of neural networks, enabling compact representation of polynomial, power-law, and rational relationships. As a result, PU-based architectures have been shown to improve extrapolation in scientific prediction tasks such as image classification and function approximation \cite{dellen2019function,dellen2024predicting}.

More recently, complex-valued PU networks have been explored for signals with amplitude and phase. In Eq.~(\ref{eq:pu_real}), the complex extension combines $\log|x_i|$ with $\arg(x_i)$ in the exponent, introducing both amplitude scaling and phase rotation via the exponential mapping. This richer inductive bias has improved robustness in MRI reconstruction and nuclear-mass prediction \cite{dellen2024predicting,li2024data,li2025advancing}.

\section{Methodology}
\subsection{Baseline: Gated Recurrent Unit}
The GRU is a recurrent neural network variant designed to capture long-term dependencies by mitigating the vanishing gradient problem. Two gating mechanisms are introduced: an update gate and a reset gate, which regulate the information flow and memory update within each recurrent unit.

Given an input vector $x_t \in \mathbb{R}^{d}$ and the previous hidden state $h_{t-1} \in \mathbb{R}^{H}$ at time step $t$, a GRU computes the update gate $z_t \in \mathbb{R}^{H}$, reset gate $r_t \in \mathbb{R}^{H}$, candidate hidden state $\tilde{h}_t \in \mathbb{R}^{H}$, and the new hidden state $h_t \in \mathbb{R}^{H}$ as follows:
\begin{align}
z_t &= \sigma(W_z x_t + U_z h_{t-1}) && \text{update gate} \label{eq:gru_update}\\
r_t &= \sigma(W_r x_t + U_r h_{t-1}) && \text{reset gate} \\
\tilde{h}_t &= \tanh(W_h x_t + U_h (r_t \odot h_{t-1})) && \text{candidate hidden state} \label{eq:gru_cand}\\
h_t &= (1 - z_t) \odot h_{t-1} + z_t \odot \tilde{h}_t && \text{new hidden state}. \label{eq:gru_hidden}
\end{align}
Here, $\sigma(\cdot)$ denotes the sigmoid activation function, $\tanh(\cdot)$ is the hyperbolic tangent, and $\odot$ represents element-wise multiplication. The learnable parameters satisfy $W_{\{z,r,h\}}\in\mathbb{R}^{H\times d}$ and $U_{\{z,r,h\}}\in\mathbb{R}^{H\times H}$. Bias terms are used in our implementation but omitted for notational simplicity.

\begin{figure}[b]
    \centering
    \includegraphics[width=0.63\linewidth]{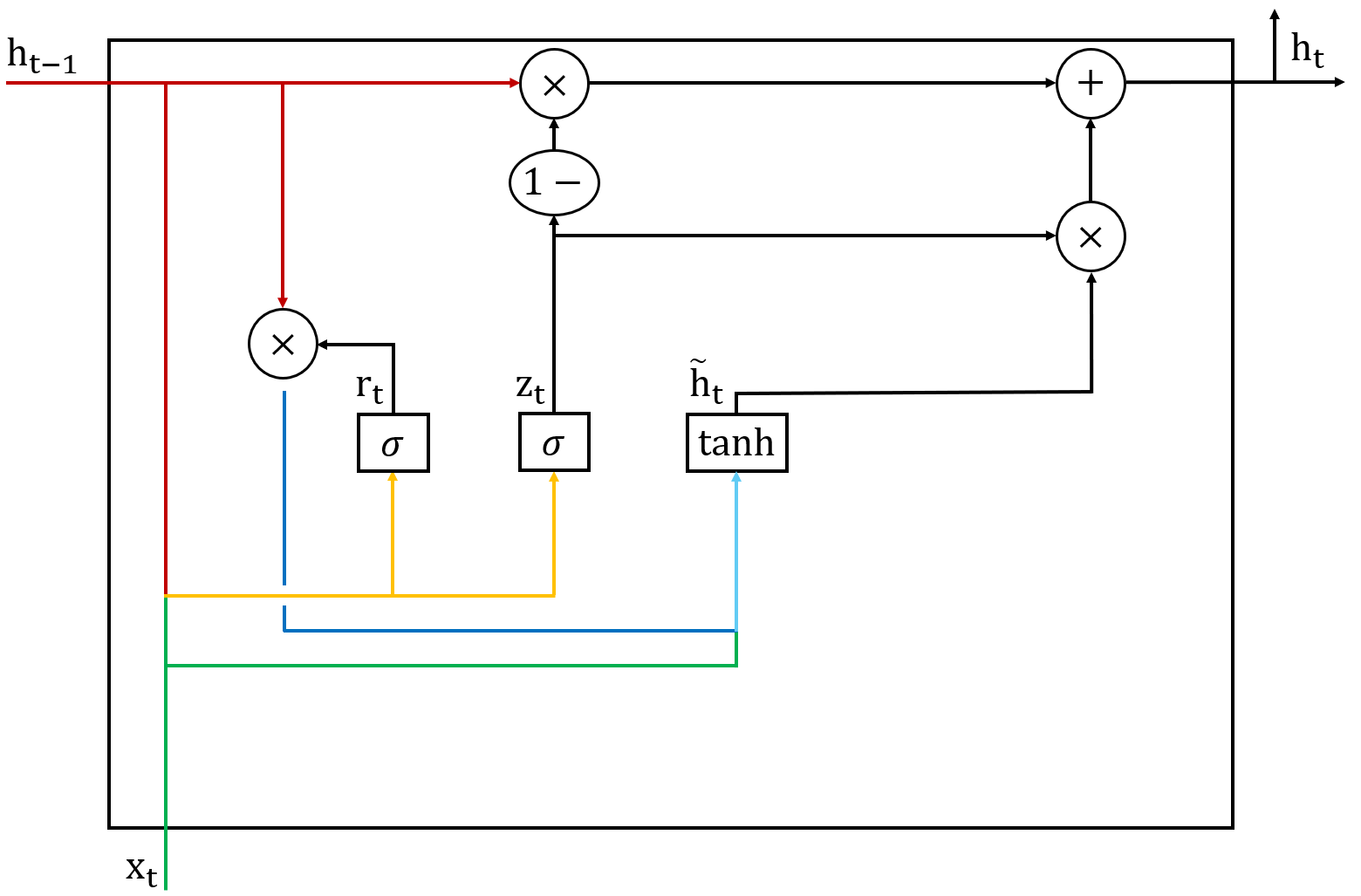}
    \vspace{-5pt}
    \caption{Computational graph of a standard GRU cell.}
    \label{fig:gru-cell}
\end{figure}
Figure~\ref{fig:gru-cell} illustrates the internal structure of a standard GRU cell. The diagram is consistent with the above equations: The input $x_t$ and previous hidden state $h_{t-1}$ are jointly used to compute the reset gate $r_t$ and update gate $z_t$. The reset gate $r_t$ controls how much of $h_{t-1}$ contributes to the candidate activation $\tilde{h}_t$. The final hidden state $h_t$ is computed as a convex combination of $h_{t-1}$ and $\tilde{h}_t$, governed by the update gate $z_t$. The parameters of the model are defined through the weight matrices associated with the update gate, reset gate, and candidate hidden state.
%BBB:  Den Satz habe ich eingefügt. Korrekt?
%Li: Yes, thank you. I slightly rephrased the sentence to avoid notation ambiguity.

\subsection{Multiplicative-Interaction Product-Unit GRU (MI-PU-GRU)}
\begin{figure}[b]
\centering
\includegraphics[width=0.63\linewidth]{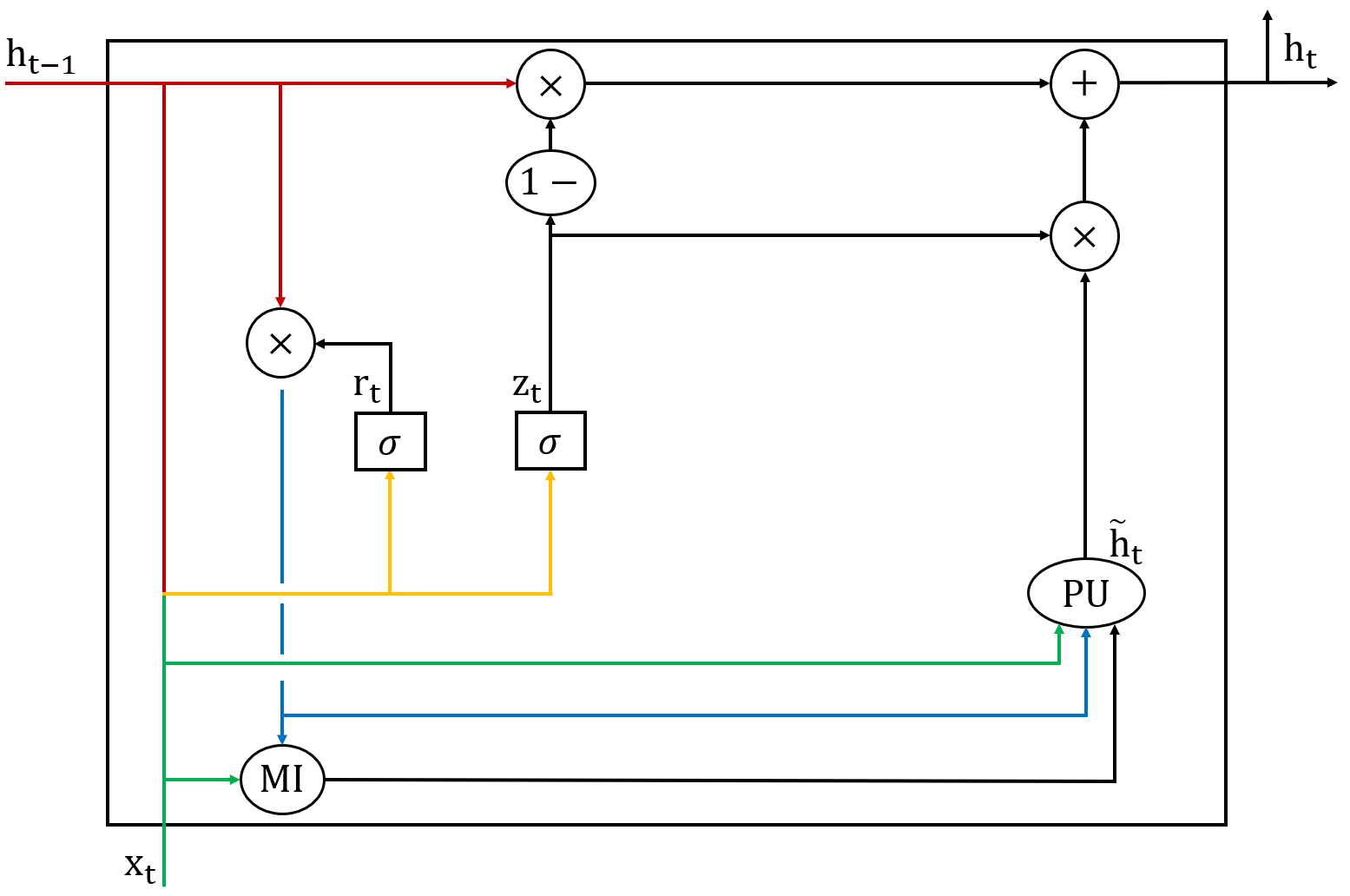}
\vspace{-5pt}
\caption{Computational graph of the MI-PU-GRU cell.}
\label{fig:mi-pu-gru}
\end{figure}

To enhance the nonlinear modeling capacity of GRUs, we introduce the MI-PU-GRU cell, which incorporates two major innovations: a multiplicative interaction (MI) branch and a product-unit (PU) transformation in the candidate-state computation. Figure~\ref{fig:mi-pu-gru} shows the MI-PU-GRU computational graph.

The MI-PU-GRU case maintains the standard GRU gating structure but modifies the candidate hidden state $\tilde{h}_t$ as follows:
\begin{align}
z_t &= \sigma(W_z x_t + U_z h_{t-1}) && \text{update gate} \\
r_t &= \sigma(W_r x_t + U_r h_{t-1}) && \text{reset gate} \\
\text{MI}_t &= \text{MI}(x_t, r_t \odot h_{t-1}) && \text{multiplicative interaction} \\
\tilde{h}_t &= \text{PU}([x_t, r_t \odot h_{t-1}, \text{MI}_t]) && \text{PU-transformed candidate} \\
h_t &= (1 - z_t) \odot h_{t-1} + z_t \odot \tilde{h}_t && \text{new hidden state}.
\end{align}
The MI term is computed using an element-wise exponential interaction between the input and hidden representations:
\begin{equation}
\text{MI}(x, h) = \exp\left( (W_{\text{mi}} x) \odot (U_{\text{mi}} h) + b_{\text{mi}} \right),
\end{equation}
where $W_{\text{mi}}$ and $U_{\text{mi}}$ are learnable projection matrices and $b_{\text{mi}}$ is a bias term. The MI branch computes a feature-wise exponential interaction between the input $x_t$ and the reset-gated previous state $r_t \odot h_{t-1}$, producing a multiplicative representation $\text{MI}_t$. The exponential function ensures that the output is strictly positive, which is necessary for the subsequent PU transformation. This representation, along with $x_t$ and $r_t \odot h_{t-1}$, is fed into a Product-Unit (PU) transformation:
\begin{equation}
\begin{split}
\mathrm{PU}(x) = \exp\Big(W_{\mathrm{pu}} \cdot\log\!\big(\max(x,\, \mathrm{softplus}(\theta) + 10^{-7})\big) + b_{\mathrm{pu}}\Big),
\end{split}
\end{equation}
where $\theta$ is a learnable threshold that ensures numerical stability of the logarithm. 

Compared with standard GRUs, MI-PU-GRU significantly enhances model expressivity by introducing higher-order multiplicative feature interactions and a PU-based nonlinear transformation.

\subsection{Additive-Multiplicative Product-Unit GRU (AM-PU-GRU)}
\begin{figure}[b]
 \centering
 \includegraphics[width=0.63\linewidth]{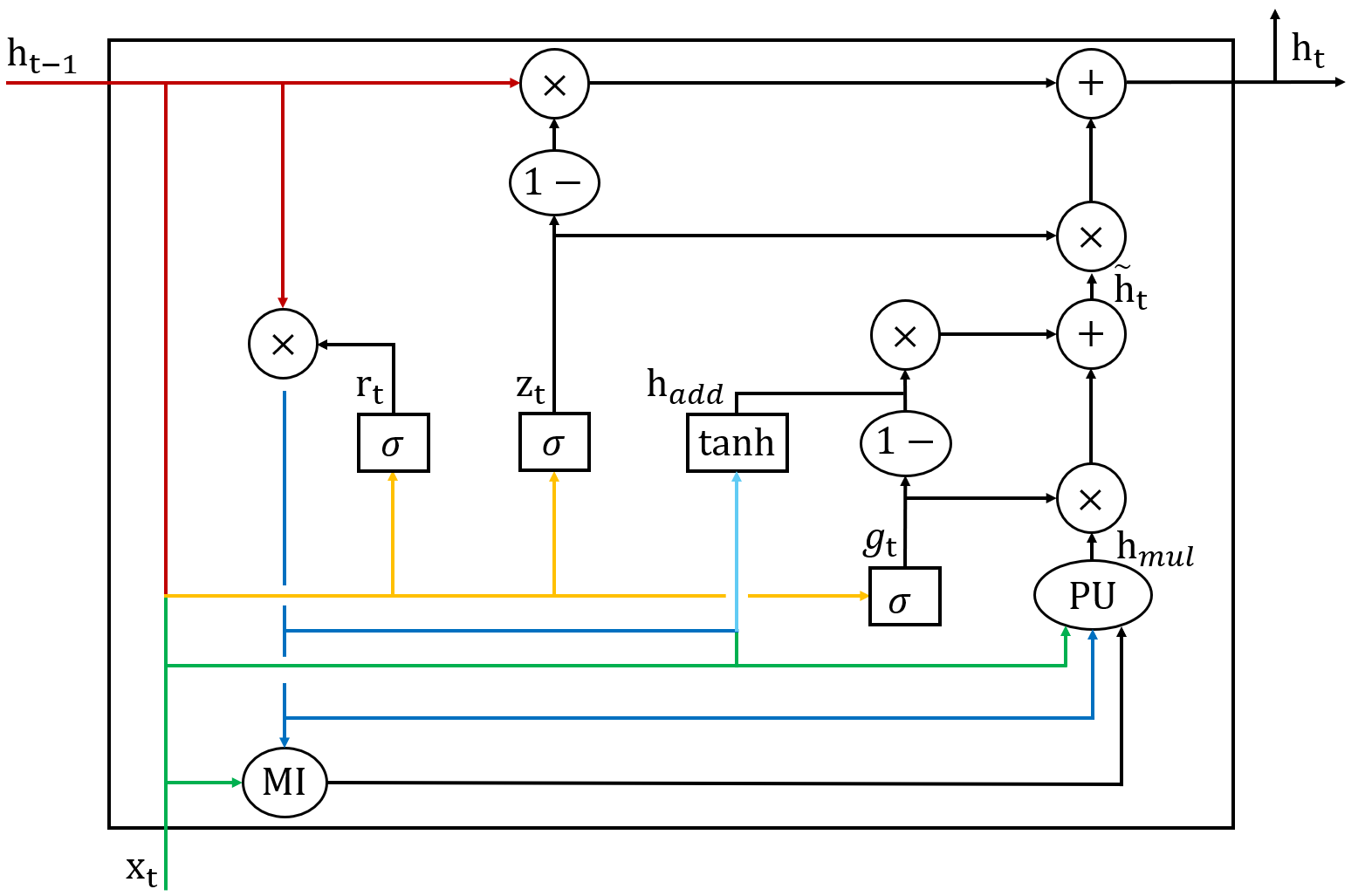}
 \vspace{-5pt}
 \caption{Computational graph of the AM-PU-GRU cell.}
 \label{fig:am-pu-gru}
\end{figure}

The AM-PU-GRU architecture further extends the MI-PU-GRU strategy by explicitly modeling both additive and multiplicative candidate paths. It introduces a learnable fusion gate to adaptively combine the two contributions, thereby unifying the strengths of traditional GRU-style linearity and PU-based multiplicative expressiveness. Figure~\ref{fig:am-pu-gru} shows the two-path candidate computation and fusion. The overall update equations are as follows:
\begin{align}
z_t &= \sigma(W_z x_t + U_z h_{t-1}) && \text{update gate} \\
r_t &= \sigma(W_r x_t + U_r h_{t-1}) && \text{reset gate} \\
g_t &= \sigma(W_g x_t + U_g h_{t-1}) && \text{fusion gate} \\
h_{\text{add}} &= \tanh(W_{\text{add}} x_t + U_{\text{add}} (r_t \odot h_{t-1})) && \text{additive candidate} \\
\text{MI}_t &= \text{MI}(x_t, r_t \odot h_{t-1}) && \text{multiplicative interaction} \\
h_{\text{mul}} &= \text{PU}([x_t, r_t \odot h_{t-1}, \text{MI}_t]) && \text{multiplicative candidate} \\
\tilde{h}_t &= g_t \odot h_{\text{mul}} + (1 - g_t) \odot h_{\text{add}} && \text{fused candidate state} \\
h_t &= (1 - z_t) \odot h_{t-1} + z_t \odot \tilde{h}_t && \text{new hidden state}.
\end{align}
Here, the additive candidate path retains a GRU-style formulation, while the multiplicative candidate is constructed using the MI and PU modules. The fusion gate $g_t$ dynamically balances the two components. This gating mechanism enables the model to flexibly adapt between linear and nonlinear dynamics depending on the temporal context.

Compared to MI-PU-GRU, the AM-PU-GRU architecture introduces an additional level of flexibility and interpretability by learning whether additive or multiplicative dynamics are more appropriate at each time step.

\subsection{Complex-Valued MI-PU-GRU}
To extend MI-PU-GRU to the complex domain, we adopt a fully complex-valued formulation. All learnable weights, hidden states, and intermediate activations are complex-valued. The real-valued input $x_t \in \mathbb{R}^{d}$ is embedded into the complex vector space via a zero-imaginary extension $\tilde{x}_t = x_t + \mathrm{i}0 \in \mathbb{C}^{d}$, and the hidden state satisfies $h_{t-1} \in \mathbb{C}^{H}$.

The update and reset gates are kept real-valued to ensure stable and interpretable gating; specifically, we apply the sigmoid to the real part of complex pre-activations, yielding $z_t, r_t \in \mathbb{R}^{H}$:
\begin{align}
z_t &= \sigma\!\left(\Re\!\left(W_z \tilde{x}_t + U_z h_{t-1}\right)\right), \\
r_t &= \sigma\!\left(\Re\!\left(W_r \tilde{x}_t + U_r h_{t-1}\right)\right), \\
\mathrm{MI}_t &= \mathrm{MI}_{\mathbb{C}}\!\left(\tilde{x}_t,\, r_t \odot h_{t-1}\right), \\
\tilde{h}_t &= \mathrm{PU}_{\mathbb{C}}\!\left([\tilde{x}_t,\, r_t \odot h_{t-1},\, \mathrm{MI}_t]\right), \\
h_t &= (1 - z_t) \odot h_{t-1} + z_t \odot \tilde{h}_t .
\end{align}
Using real-valued gates (bounded in $[0,1]$) provides a numerically stable and interpretable mixing of complex hidden states without introducing additional phase rotations.

The MI branch performs feature-wise exponential interactions in the complex domain:
\begin{equation}
\mathrm{MI}_{\mathbb{C}}(x, h) =
\exp\!\left( (W_{\mathrm{mi}} x) \odot (U_{\mathrm{mi}} h) + b_{\mathrm{mi}} \right),
\end{equation}
where all parameters are complex-valued. Since $\exp(a+\mathrm{i}b)=\exp(a)\big(\cos b + \mathrm{i}\sin b\big)$, the interaction introduces both amplitude scaling and phase rotation, enhancing expressiveness.

The PU transformation in the complex domain is defined as
\begin{equation}
\mathrm{PU}_{\mathbb{C}}(x)=
\exp\!\left(W_{\mathrm{pu}}\big(\log r(x)+\mathrm{i}\,\phi(x)\big)+b_{\mathrm{pu}}\right),
\end{equation}
where $r(x)=\max\!\big(|x|,\mathrm{softplus}(\theta)+\epsilon\big)$, $\phi(x)=\arg(x)$, and $\epsilon=10^{-7}$. Here, the logarithm is applied to the stabilized magnitude $r(x)$ and the phase term $\phi(x)$ is handled explicitly.

Since our downstream task requires real-valued predictions, the final output of the complex-valued recurrent network is mapped back to the real domain by extracting the real part of the hidden state. This mapping ensures compatibility with standard loss functions defined over $\mathbb{R}$.

\subsection{Complex-Valued AM-PU-GRU}
As with standard AM-PU-GRU, complex AM-PU-GRU builds upon complex MI-PU-GRU by integrating an additive path and a fusion gate to balance linear and nonlinear dynamics in $\mathbb{C}$. The update procedure is given by:
\begin{align}
z_t &= \sigma\!\left(\Re\!\left(W_z \tilde{x}_t + U_z h_{t-1}\right)\right), \\
r_t &= \sigma\!\left(\Re\!\left(W_r \tilde{x}_t + U_r h_{t-1}\right)\right), \\
g_t &= \sigma\!\left(\Re\!\left(W_g \tilde{x}_t + U_g h_{t-1}\right)\right), \\
h_{\mathrm{add}} &= \tanh\!\left(\Re\!\left(W_{\mathrm{add}} \tilde{x}_t + U_{\mathrm{add}}\left(r_t \odot h_{t-1}\right)\right)\right), \\
\mathrm{MI}_t &= \mathrm{MI}_{\mathbb{C}}\!\left(\tilde{x}_t,\, r_t \odot h_{t-1}\right), \\
h_{\mathrm{mul}} &= \mathrm{PU}_{\mathbb{C}}\!\left([\tilde{x}_t,\, r_t \odot h_{t-1},\, \mathrm{MI}_t]\right), \\
\tilde{h}^{\mathrm{add}}_t &= h_{\mathrm{add}} + \mathrm{i}0, \\
\tilde{h}_t &= g_t \odot h_{\mathrm{mul}} + (1 - g_t) \odot \tilde{h}^{\mathrm{add}}_t, \\
h_t &= (1 - z_t) \odot h_{t-1} + z_t \odot \tilde{h}_t .
\end{align}
Here, $h_{\mathrm{add}} \in \mathbb{R}^{H}$ is recast to $\mathbb{C}^{H}$ before fusion. The use of real-valued gates ensures numerically stable and interpretable mixing while allowing the hidden state to evolve in the full complex plane.
 
\subsection{Task Formulation}
Our goal is to predict the mass excess of an atomic nucleus from sequential nuclear structure data, which we formulate as a sequence-to-one regression problem. For each target nucleus with proton number $Z^\star$ and neutron number $N^\star$, we construct a short ordered sequence of $T=5$ neighboring nuclides based on proximity in the $(Z,N)$ chart.

For a target nucleus $(Z^\star,N^\star)$ with mass number $A^\star=Z^\star+N^\star$, the first $T-1=4$ elements are selected from the corresponding reference set of experimentally measured nuclei. Specifically, excluding the target nucleus itself, we consider nuclei satisfying $A\leq A^\star$ and select the four nearest ones in the $(Z,N)$ plane as context nuclei; their measured mass excess values are used as input features. The final element $x_T$ corresponds to the target nucleus; its mass excess value in the input is initialized by the WS4 mass model (i.e., $\mathrm{ME}_T=\mathrm{ME}^{\mathrm{WS4}}(Z^\star,N^\star)$) to provide a physically informed prior.

Given $X$, the model learns a nonlinear mapping $f:\mathbb{R}^{T\times 3}\rightarrow \mathbb{R}$ to predict the ground-truth mass excess $y=\mathrm{ME}^{\mathrm{AME}}(Z^\star,N^\star)$ of the target nucleus. This design allows the model to exploit local continuity over neighboring nuclides while combining empirical measurements and theoretical priors.

To capture the complex nonlinear dependencies inherent in nuclear-mass systematics, we propose complex-valued extensions of the MI-PU-GRU and AM-PU-GRU architectures, which integrate multiplicative interactions and product-unit transformations into the recurrent framework.

In addition to these primary models, we also implement their real-valued counterparts (MI-PU-GRU and AM-PU-GRU) to provide a controlled comparison between real and complex formulations. Furthermore, we design ablation variants in the complex domain, including PU-GRU, MI-GRU, AM-GRU, and the baseline GRU, by selectively disabling specific architectural components. These ablations allow us to isolate and quantify the contribution of multiplicative interactions, product-unit transformations, and additive–multiplicative fusion. Finally, to evaluate robustness with respect to theoretical priors, we perform prior stability experiments where the target nucleus is initialized using the semi-empirical mass formula (SEMF), which incorporates volume, surface, Coulomb, asymmetry, and pairing terms. The SEMF-based mass excess estimate is computed using a simplified version of the Weizsäcker formula \cite{weizsacker1935theorie}:
\begin{equation}
ME = (Z M_p + N M_n + Z M_e - B - A) \cdot 931.5 \times 10^3 \text{ keV},
\end{equation}
where $B$ is the total binding energy, which is calculated as
\begin{equation}
B = a_v A - a_s A^{2/3} - a_c \frac{Z(Z-1)}{A^{1/3}} - a_a \frac{(A - 2Z)^2}{A} + \delta,
\end{equation}
with $A=Z+N$ and the pairing term $\delta$ defined as $\delta=+a_p/\sqrt{A}$ for even--even nuclei, $\delta=-a_p/\sqrt{A}$ for odd--odd nuclei, and $\delta=0$ otherwise.

\section{Experiments and Results}

\subsection{Experimental Setup}

\paragraph{Experiment I: Interpolation.}
This experiment assesses the model's interpolation capability within the domain of experimentally known nuclides. We randomly split the AME2020 dataset into training, validation, and test subsets using a 70\%-15\%-15\% ratio. During sequence construction for the validation and test sets, only the experimentally measured mass-excess values from the training set are used to provide input features for precursor nuclei. That is, all non-target entries in a sequence are filled using training set information, ensuring no data leakage. Model selection is based on validation performance: the version of the model that achieves the lowest validation loss during training is retained and used for final evaluation on the test set.

\paragraph{Experiment II: Temporal Extrapolation.}
To test the extrapolation capability of our models, we simulate a temporal generalization scenario by training on AME2016-era nuclides and testing on nuclides added in the subsequent AME2020 release. This setup allows us to evaluate the model's ability to generalize to newly measured nuclei that were previously unknown. In addition, we reserve 5\% of the AME2016 data as an interpolation test set to provide a comprehensive assessment of the model's performance on both interpolation and extrapolation tasks within the same experimental framework. As in Experiment I, all sequences for both extrapolation and interpolation test sets are constructed using only nuclei from the training set to provide precursor mass excess values.

\paragraph{Training Setup.}
All models are trained in PyTorch for 3000 epochs using MSE loss and the RAdam optimizer (initial learning rate $10^{-3}$). We use a step learning-rate schedule that halves the learning rate every 500 epochs, and set the batch size to 64.

\subsection{Main Results}
To evaluate the predictive capability of the proposed architectures, we first compare the baseline real-valued GRU with the complex-valued MI-PU-GRU and AM-PU-GRU under the WS4 prior. For a fair comparison, we control the parameter scale by adjusting network depth: the baseline GRU is implemented with 120 stacked recurrent layers, while the complex MI-PU-GRU and AM-PU-GRU are implemented with 60 and 50 layers, respectively. This configuration results in a comparable number of trainable parameters across all models. For complex-valued architectures, each learnable weight is represented as $a+ib$, where $a$ is the real and $b$ the imaginary part of the complex weight, and thus counts as two trainable parameters when reporting parameter sizes.

Each network is trained independently five times without fixing the random seed, in order to account for stochasticity in initialization and training dynamics. We report the mean and standard deviation of the RMSE over these runs. 

The interpolation results of Experiment I are summarized in Table~\ref{tab:main_e1}. As can be seen, both complex-valued PU-GRU variants consistently outperform the real-valued GRU baseline. In particular, the complex AM-PU-GRU model achieves the lowest interpolation error (0.227 MeV) and the lowest sample standard deviation (0.004 MeV), highlighting the effectiveness of combining additive and multiplicative candidate paths within the complex domain.
\begin{table}[ht]
\centering
\setlength{\tabcolsep}{10pt}
\caption{Principal results of \textbf{Experiment I} using WS4 estimates. Here, RV denotes the real-valued model, CV the complex-valued model. Results are reported as the mean $\pm$ sample standard deviation (MeV). $N_{\mathrm{param}}$ indicates the number of trainable parameters. IntRMSE denotes interpolation RMSE (and ExtRMSE denotes extrapolation RMSE, when applicable). The best result in each column is highlighted in bold.}
\begin{tabular}{lcc}
\hline
Model         & $N_{\mathrm{param}}$      & IntRMSE (MeV)                   \\ \hline
RV GRU        & 45121                   & 0.242 ± \textbf{0.004}          \\
CV MI-PU-GRU  & 38165                   & 0.236 ± 0.005                   \\
CV AM-PU-GRU  & 37605                   & \textbf{0.227} ± \textbf{0.004} \\ \hline
\end{tabular}
\label{tab:main_e1}
\end{table}

The results of Experiment II are summarized in Table~\ref{tab:main_e2}. Compared to the real-valued GRU baseline, both complex PU-GRU variants yield lower extrapolation errors, with the complex AM-PU-GRU again achieving the best overall performance. In particular, AM-PU-GRU attains the lowest interpolation RMSE of 0.253 ± 0.003 MeV and the lowest extrapolation RMSE of 0.179 ± 0.015 MeV. These findings indicate that the proposed complex architectures improve both interpolation and temporal-extrapolation performance under the present experimental setup. In particular, the extrapolation capability of the complex AM-PU-GRU is significantly stronger, indicating its effectiveness in capturing long-range structural dependencies in nuclear mass systematics. Moreover, AM-PU-GRU exhibits the most stable extrapolation performance, as reflected by the lowest standard deviation among all models.
\begin{table}[ht]
\centering
\setlength{\tabcolsep}{10pt}
\caption{Main results of \textbf{Experiment II} using WS4 estimates.}
\begin{tabular}{lcc}
\hline
Model        & IntRMSE (MeV)                   & ExtRMSE (MeV)                   \\ \hline
RV GRU       & 0.261 ± \textbf{0.003}          & 0.205 ± 0.018                   \\
CV MI-PU-GRU & 0.254 ± \textbf{0.003}          & 0.186 ± 0.020                   \\
CV AM-PU-GRU & \textbf{0.253} ± \textbf{0.003} & \textbf{0.179} ± \textbf{0.015} \\ \hline
\end{tabular}
\label{tab:main_e2}
\end{table}

The lower extrapolation RMSE in Table~2 does not mean that extrapolation is generally easier. In Experiment II, the interpolation set is a 5\% hold-out from AME2016, whereas the extrapolation set consists of nuclei newly added in AME2020, so the two test sets differ in composition. Since prediction depends on local neighborhood structure in the $(Z,N)$ chart and the WS4 prior, some AME2020-added nuclei may be easier to predict than the held-out AME2016 subset.

We further compare our best-performing models with existing approaches from the literature, as summarized in the Discussion section.

\subsection{Comparative Analysis: Real vs. Complex}
To assess the effect of complex-valued modeling, we compare the performance of real-valued PU-GRU variants with their complex-valued counterparts under the WS4 prior. Specifically, we implement a 90-layer real-valued MI-PU-GRU and a 75-layer real-valued AM-PU-GRU, such that their parameter counts are comparable to those of the 60-layer complex MI-PU-GRU and the 50-layer complex AM-PU-GRU, respectively. Each real-valued network is trained independently three times, and the reported performance is the average RMSE across runs.

The results, summarized in Tables~\ref{tab:comparative_e1} and~\ref{tab:comparative_e2}, indicate that the complex-valued PU-GRU variants consistently outperform their real-valued counterparts in both interpolation and extrapolation settings. This demonstrates that extending PU-GRU architectures into the complex domain yields richer representational capacity by jointly modeling amplitude and phase dynamics, leading to improved predictive accuracy.
\begin{table}[ht]
\centering
\setlength{\tabcolsep}{10pt}
\caption{Results of \textbf{Experiment I} for real-valued variants.}

%JWC:in this table, headers for parameters and interpolated RMSE are %unsatisfactory. Introduce and define labels, e.g. IntRMSE.  Also, the %brackets around MeV in the table below are strange.  Corresponding comments %apply to next table, including Interp. and Extrap. These things could be %addressed in the captions. Similar comments apply to later tables.
%Li: I updated the table headers to use clearly defined labels and standardized the unit formatting to “(MeV)” throughout.

\begin{tabular}{lcc}
\hline
Model        & $N_{\mathrm{param}}$ & IntRMSE (MeV) \\ \hline
RV MI-PU-GRU & 42580              & 0.240          \\
RV AM-PU-GRU & 41710              & 0.233           \\ \hline
\end{tabular}
\label{tab:comparative_e1}
\end{table}
\begin{table}[ht]
\centering
\setlength{\tabcolsep}{10pt}
\caption{Results of \textbf{Experiment II} for real-valued variants.}
\begin{tabular}{lcc}
\hline
Model        & IntRMSE (MeV)      & ExtRMSE (MeV)      \\ \hline
RV MI-PU-GRU & 0.255              & 0.196              \\
RV AM-PU-GRU & 0.258              & 0.194              \\ \hline
\end{tabular}
\label{tab:comparative_e2}
\end{table}

\subsection{Ablation Study}

To isolate the contributions of individual architectural components, we have performed an ablation study in the complex domain. We implement four variants: a complex-valued GRU with 80 layers, a 70-layer complex MI-GRU (without PU and additive path), a 60-layer complex AM-GRU (without PU but with additive multiplicative fusion), and an 80-layer complex PU-GRU (retaining PU but without explicit MI or AM mechanisms). All models are configured to have comparable parameter counts for a fair comparison.

%%Great paragraph!JWC
The results are presented in Tables~\ref{tab:ablation_e1} and~\ref{tab:ablation_e2}. We observe that neither the complex MI-GRU nor the complex AM-GRU improves upon the baseline complex GRU, and in fact both exhibit degraded performance. By contrast, the complex PU-GRU model achieves a substantial reduction in error, clearly outperforming the baseline GRU. This indicates that the product-unit transformation is the key factor driving improvements, while multiplicative interactions or additive–multiplicative fusion alone are insufficient to enhance predictive accuracy.
\begin{table}[ht]
\centering
\setlength{\tabcolsep}{10pt}
\caption{Ablation results of \textbf{Experiment I} with complex-valued GRU, MI-GRU, AM-GRU, and PU-GRU implementation.}
\begin{tabular}{lcc}
\hline
Model     & $N_{\mathrm{param}}$ & IntRMSE (MeV) \\ \hline
CV GRU    & 40482              & 0.400         \\
CV MI-GRU & 41582              & 0.801         \\
CV AM-GRU & 38522              & 2.051         \\
CV PU-GRU & 40482              & 0.261         \\ \hline
\end{tabular}
\label{tab:ablation_e1}
\end{table}
\begin{table}[ht]
\centering
\setlength{\tabcolsep}{10pt}
\caption{Ablation results of \textbf{Experiment II} with complex-valued GRU, MI-GRU, AM-GRU and PU-GRU strategies.}
\begin{tabular}{lcc}
\hline
Model     & IntRMSE (MeV)      & ExtRMSE (MeV)      \\ \hline
CV GRU    & 0.493              & 0.259 \\
CV MI-GRU & 0.843              & 0.241 \\
CV AM-GRU & 1.316              & 0.282 \\
CV PU-GRU & 0.261              & 0.197 \\ \hline
\end{tabular}
\label{tab:ablation_e2}
\end{table}

\subsection{Prior Stability Analysis}
Finally, we evaluate robustness to the choice of the theoretical baseline estimate used to initialize the target nucleus (WS4 vs. SEMF). In this experiment, the network configurations remain the same as in the main results, i.e., a 120-layer real-valued GRU baseline, a 60-layer complex MI-PU-GRU, and a 50-layer complex AM-PU-GRU. The only difference lies in the initialization of the target nucleus, 
where we replace the WS4 estimate with the SEMF, which incorporates volume, surface, Coulomb, asymmetry, and pairing terms.
%JWC: did you define SEMF above; otherwise write it out.
%Li: yes, in 3.6

The results are summarized in Tables~\ref{tab:prior_e1} and~\ref{tab:prior_e2}. Compared to the WS4-based experiments, the relative ranking of models is unchanged: both complex PU-GRU variants outperform the real-valued GRU, and the complex AM-PU-GRU version continues to achieve the best overall accuracy. These findings confirm that the proposed architectures are robust to the choice of baseline initialization model (WS4 vs. SEMF) and maintain stable capability under this change.
\begin{table}[ht]
\centering
\setlength{\tabcolsep}{10pt}
%JWC comment!!: not clear by what you mean by "Prior" here. Or 1G/rio.  I may %have disturbed or even eliminated something the text in this region of the %text. But it is dangerous to search now.  My control of this file on this %ancient laptop is fragile.  And editing text without line breaks gets to be
%very tricky.
%Li: I revised the preceding text to clarify that “prior” refers specifically to the theoretical baseline used to initialize the target nucleus (WS4 vs. SEMF).
\caption{Prior stability results of \textbf{Experiment I} under SEMF initialization.}
\begin{tabular}{lc}
\hline
%JWC meaning of Interp. may not be clear.  Define a IRMSE or IntRMSE?
%Also, what's going on with the symbols enclosing MeV below?
%Li: This may be due to different editors. My original notation was [MeV], but it has now been changed to (MeV).
Model        & IntRMSE (MeV)          \\ \hline
RV GRU       & 0.414                           \\
CV MI-PU-GRU & 0.402                           \\
CV AM-PU-GRU & \textbf{0.401}                  \\ \hline
\end{tabular}
\label{tab:prior_e1}
\end{table}
\begin{table}[ht]
\centering
\setlength{\tabcolsep}{10pt}
\caption{Prior stability results of \textbf{Experiment II} under SEMF initialization.}
\begin{tabular}{lcc}
\hline
Model        & IntRMSE (MeV)      & ExtRMSE (MeV)      \\ \hline
RV GRU       & 0.481 & 0.598 \\
CV MI-PU-GRU & 0.460 & 0.505 \\
CV AM-PU-GRU & 0.455 & \textbf{0.477} \\ \hline
\end{tabular}
\label{tab:prior_e2}
\end{table}

\section{Discussion}
Beyond comparisons among our proposed variants, it is important to contextualize the performance of the PU-GRU architectures with respect to existing approaches in the literature. Numerous machine-learning methods have been previously applied to nuclear mass prediction, including MDN, categorical
%JWC: have you defined MDN??? 
%Li:yes, in 1
gradient boosting trees (CatBoost), and fully connected neural networks (FCNN). We summarize representative results reported in prior studies and compare them against our best-performing model, the complex AM-PU-GRU.
Table~\ref{tab:lit_t1} summarizes the interpolation performance of various machine learning models and our models on the AME2020 dataset (Experiment I). All models are evaluated on the task of predicting nuclear mass excess values using various input feature sets and network architectures. 
\begin{table}[ht]
\centering
\caption{Comparison with existing approaches in \textbf{Experiment I}. All RMSE values are reported in MeV. The column “Selection criterion” specifies the nuclide selection rule, $N_{\mathrm{feat}}$ denotes the number of input features per nuclide, and $E_{\mathrm{measure}}$ refers to the experimental uncertainty of the measured mass-excess values.}
\setlength{\tabcolsep}{10pt}
\begin{tabular}{lccc}
\hline
Model     & Selection criterion  & $N_{\mathrm{feat}}$ & IntRMSE (MeV) \\ \hline
XGBoost \cite{pandey2024prediction}       &                                                                                   &            & 3.702                                                                     \\
MLP Regressor \cite{pandey2024prediction} &                                                                                   & 4          & 3.128                                                                     \\
RFR \cite{pandey2024prediction}           &                                                                                   &            & 3.089                                                                     \\ \hline
MISR \cite{munoz2025discovering}          & $12 \le Z \le 50$                                                               & 10         & 0.99                                                                      \\ \hline
MDN \cite{mumpower2023bayesian}           & $Z, N \ge 5$                                                                      & 9          & 0.395                                                                     \\ \hline
MDN \cite{li2024atomic}                   & $Z \ge 10$                                                                        & 11         & 0.246                                                                     \\ \hline
CatBoost \cite{guo2025probing}            & \begin{tabular}[c]{@{}c@{}}$8 \le Z \le 108$\\ $8 \le N \le 162$\end{tabular} & 7          & 0.189                                                                     \\ \hline
PI-FCNN \cite{huang2025validation}        & \begin{tabular}[c]{@{}c@{}}$Z, N > 20$\\ $E_{measure}< 100 \,keV$\end{tabular}   & 13         & 0.122                                                                     \\ \hline
CPUN \cite{dellen2024predicting}          &                                                                                   & 10         & 0.438                                                                     \\ \hline
RNN \cite{jalili2025deep}                 &                                                                                   &            & 0.601                                                                     \\
LSTM \cite{jalili2025deep}                & $Z, N \ge 8$                                                                      & 11         & 0.557                                                                     \\
GRU \cite{jalili2025deep}                 &                                                                                   &            & 0.459                                                                     \\ \hline
CV AM-PU-GRU {[}ours{]}                   & $Z, N \ge 8$                                                                      & 3          & 0.227                                                                     \\ \hline
\end{tabular}
\label{tab:lit_t1}
\end{table}

As shown in Table~\ref{tab:lit_t1}, Jalili et al. \cite{jalili2025deep} reported results for several recurrent architectures, including GRU, RNN, and LSTM, but their models utilize broader input features, as well as standard recurrent designs without product-unit augmentation. Our complex-valued AM-PU-GRU model achieves superior performance with a more compact and physics-informed input representation, achieving an RMSE of 0.227 MeV for mass-excess prediction.

We also compare with the CPUN proposed by Dellen et al. \cite{dellen2024predicting}, which yields an RMSE of 0.438 MeV. While their model also incorporates product units and complex-valued representations, it is based on a fully connected feedforward architecture and lacks the temporal modeling capability of our recurrent design. In contrast, our PU-GRU framework integrates multiplicative interactions into a recurrent paradigm and explicitly exploits sequence-level dependencies among neighboring nuclides, leading to improved performance over CPUN \cite{dellen2024predicting}.

Nevertheless, our model does not outperform the best results reported in the literature, notably CatBoost \cite{guo2025probing} (0.189 MeV) and Physics-Informed FCNN \cite{huang2025validation} (0.122 MeV). This gap is partly due to differences in data selection and domain-specific constraints. For example, \cite{huang2025validation} evaluates only nuclides with $Z,N>20$ and measured error $<100$ keV, excluding many difficult edge cases, while \cite{guo2025probing} imposes upper bounds of $Z \le 108$ and $N \le 162$, which also simplifies the task. By contrast, our model is evaluated over a broader nuclide range without such filtering, making the problem more challenging. In addition, unlike tree-based methods such as CatBoost \cite{guo2025probing}, which are not end-to-end and often depend on feature engineering, our PU-GRU follows an end-to-end paradigm that maps raw nuclear sequences directly to mass excess values while retaining modular interpretability.

Table~\ref{tab:lit_t2} summarizes the interpolation performance on AME2016 and temporal extrapolation on AME2020 of various machine-learning models and our models (Experiment II). Across both interpolation and extrapolation tasks, our complex-valued AM-PU-GRU model consistently achieves the lowest prediction error (0.253 MeV and 0.179 MeV) among all compared methods. Furthermore, its performance even surpasses certain non-end-to-end approaches, such as physics-informed FCNN \cite{huang2025validation} and CNN-WS4 \cite{lu2025nuclear}, highlighting the effectiveness of learning nuclear mass systematics directly from sequential data.
\begin{table*}[htbp]
\centering
\caption{Comparison with existing approaches in \textbf{Experiment II}. All RMSE values are reported in MeV.}
\setlength{\tabcolsep}{6pt}
\begin{tabular}{lcccc}
\hline
Model     & Selection criterion  & $N_{\mathrm{feat}}$ & IntRMSE & ExtRMSE\\ \hline
GPR \cite{yuksel2024nuclear} & $Z, N \ge 8$ & 12 & 0.26 & 0.67 \\
SVR \cite{yuksel2024nuclear} & & & 0.39 & 0.74 \\ \hline
MDN \cite{mumpower2022physically} & $Z \ge 20$ & 8 & 0.316 & 0.336 \\ \hline
CNN-WS4 \cite{lu2025nuclear} & $Z, N > 8$ & 9 & - & 0.211 \\ \hline
PI-FCNN \cite{huang2025validation} & \begin{tabular}[c]{@{}c@{}}$Z, N > 20$ and \\ $E_{measure}< 100\, keV$\end{tabular} & 13 & - & 0.191 \\ \hline
CV AM-PU-GRU [ours] & $Z, N \ge 8$ & 3 & 0.253 & 0.179\\ \hline
\end{tabular}
\label{tab:lit_t2}
\end{table*}

\section{Conclusion}
We have presented complex-valued extensions of the MI-PU-GRU and AM-PU-GRU architectures for nuclear mass prediction. By embedding multiplicative interactions and product-unit transformations into recurrent units and extending them to the complex domain, our models capture nonlinear dependencies and long-range structural correlations that are difficult to represent with standard additive recurrent architectures.

Through systematic evaluation on interpolation and extrapolation tasks using AME2016 and AME2020 datasets, we have shown that the complex AM-PU-GRU model achieves the best overall performance, consistently yielding the lowest RMSE across all scenarios. Comparative analyses further reveal that complex-valued models outperform their real-valued counterparts, while ablation studies identify the product-unit transformation as the critical component driving accuracy improvements. Moreover, prior stability experiments confirm that the proposed architectures maintain consistent predictive capability under both WS4 and SEMF initialization, demonstrating robustness with respect to theoretical priors.

Beyond outperforming existing end-to-end machine learning approaches, our models also surpass certain non-end-to-end correction schemes, underscoring the potential of product-unit recurrent architectures as powerful tools for scientific prediction tasks. Future work will explore extending complex PU-based recurrent networks to other nuclear observables and to broader domains where extrapolation capability is essential.

\bibliographystyle{splncs04}
\bibliography{reference}

\end{document}